\documentclass[a4paper,conference]{IEEEtran}
\IEEEoverridecommandlockouts
\usepackage{cite}
\usepackage{amsmath,amssymb,amsfonts}
\usepackage{algorithmic}
\usepackage{graphicx}
\usepackage{textcomp}
\usepackage{xcolor}
\usepackage{color}
\newcommand{\argmin}{\mathop{\rm arg~min}\limits}
\usepackage{algorithm,algorithmic}
\usepackage{multirow}
\def\BibTeX{{\rm B\kern-.05em{\sc i\kern-.025em b}\kern-.08em
    T\kern-.1667em\lower.7ex\hbox{E}\kern-.125emX}}
\begin{document}

\title{Channel Planting for Deep Neural Networks using Knowledge Distillation\\
\thanks{This work was partly supported by JSPS KAKENHI Grant Number 16K00239.}
}

\author{\IEEEauthorblockN{Kakeru Mitsuno, Yuichiro Nomura and Takio Kurita}
\IEEEauthorblockA{\textit{Graduate School of Advanced Science and Engineering} \\
\textit{Hiroshima University}\\
Higashi Hiroshima, Japan \\
mitsunokakeru@gmail.com, \{d202757, tkurita\}@hiroshima-u.ac.jp
}
}

\maketitle

\begin{abstract}

In recent years, deeper and wider neural networks have shown excellent performance in computer vision tasks, while their enormous amount of parameters results in increased computational cost and overfitting.
Several methods have been proposed to compress the size of the networks without reducing network performance. 
Network pruning can reduce redundant and unnecessary parameters from a network. Knowledge distillation can transfer the knowledge of deeper and wider networks to smaller networks. 
The performance of the smaller network obtained by these methods is bounded by the predefined network. 
Neural architecture search has been proposed, which can search automatically the architecture of the networks to break the structure limitation. 
Also, there is a dynamic configuration method to train networks incrementally as sub-networks. 

In this paper, we present a novel incremental training algorithm for deep neural networks called {\em ``planting``}. 
Our planting can search the optimal network architecture with smaller number of parameters for improving the network performance by augmenting channels incrementally to layers of the initial networks while keeping the earlier trained parameters fixed.
Also, we propose using the knowledge distillation method for training the channels planted. 
By transferring the knowledge of deeper and wider networks, we can grow the networks effectively and efficiently.
We evaluate the effectiveness of the proposed method on different datasets such as CIFAR-10/100 and STL-10. 
For the STL-10 dataset, we show that we are able to achieve comparable performance with only $7\%$ parameters compared to the larger network and reduce the overfitting caused by a small amount of the data.

\end{abstract}


\section{Introduction}

Deep neural networks (DNNs) have been successful with superior performance in computer vision tasks such as image classification \cite{krizhevsky2012imagenet, szegedy2015going}. 
Meanwhile, the network becomes deeper and wider, requiring excessive amount of parameters to achieve excellent performance \cite{he2016deep,huang2017densely}, which increases the computational cost and cause overfitting. 

To improve performance while reducing the computational cost, various network pruning approaches for compressing the size of the network have been proposed. 
Network pruning can reduce unnecessary parameters while keeping network performance,
by using the Taylor expansion of the loss function \cite{lecun1990optimal, hassibi1993second}, 
enforcing unnecessary parameter to be 0 with sparse regularization \cite{han2015learning, wen2016learning, mitsuno2020filter}, 
evaluating the importance of the parameter based on the norm \cite{li2016pruning, he2018soft} 
or using the scaling parameter of batch normalization layers \cite{liu2017learning}.

There is another way to compress the size of networks, called knowledge distillation. 
Knowledge distillation is a technique for transferring the knowledge of a deeper and wider network or ensemble network (teacher networks) to smaller and shallower networks (student networks) by getting close the output of student networks to teacher networks.
To transfer the knowledge, the L2 loss or the KL-divergence is used as the loss function of the method \cite{ba2014deep, hinton2015distilling}.

However, these methods utilizing the larger predefined network with handcrafted or existing architecture has an upper bound of the network performance. 
To search automatically optimal network architecture, Neural Architecture Search (NAS) is introduced.
NAS explore the width and depth of networks efficiently for the training task,
by using a recurrent neural network as the controller \cite{zoph2016neural},
using graph-based algorithm \cite{pham2018efficient, dong2019searching},
or optimizing search space \cite{wu2019fbnet}.

An approach similar to NAS is incremental training\cite{tann2016runtime}.  
Incremental training is a dynamic configuration technique for DNNs, that initially train a subset of channels in each layer and gradually add in more channels while keeping the earlier trained channels fixed. 
By training with this method, we can obtain a flexibility in the network training, which can dynamically adjust the DNNs to reduce the computational cost as long as the accuracy of the classification results is not compromised.

The existing incremental training method uses the handcrafted network architecture as a base network and divides it into several sub-networks.
There is an upper bound of the network performance since the architecture of the sub-networks is fixed.

In this paper, we introduce a novel incremental training method for DNNs called {\em ``planting``}. 
The planting method can search the optimal network architecture for the training task with smaller parameters by planting channels incrementally to initial networks while keeping the earlier trained channels fixed for improving the network performances.
Explore the architecture of the network by planting channels in a layer where the error is reduced by adding channels.
For the training of the planted channels, the proposed method utilizes the knowledge transfer method.
The parameters in the augmented channels are trained to complement the error of the earlier trained network by imitating the behavior of the teacher network.
The contributions of this paper are summarized as follows:
\begin{itemize}
\item We propose a novel incremental training method for DNNs called {\em planting}, that can train smaller network with excellent performance and find the optimal network architecture automatically.
\item We introduce the knowledge transfer to train planted channels.
\item We have performed experiments to evaluate the effectiveness of the proposed method on different datasets (CIFAR-10, CIFAR-100, and STL-10). 
\end{itemize}

\section{Related Works}

\subsection{Network Pruning}
Network pruning can efficiently prune unnecessary weights or filters to compress DNNs, keeping network performance.
Unstructured pruning is a method to remove redundant weights of DNNs one by one.
Optimal brain damage \cite{lecun1990optimal} and optimal brain surgeon \cite{hassibi1993second} prune unimportant weights based on the Hessian matrix, which is obtained by the Taylor expansion of the loss function.
Some works \cite{molchanov2016pruning,lin2018accelerating,peng2019collaborative,molchanov2019importance} also used Taylor expansions to evaluate the influence of pruned filters to the classification loss.
Han et al. \cite{han2015learning} and Louizos et al. \cite{louizos2017learning} utilized unstructured sparse regularizations to reduce unnecessary weights.
Structured pruning is a method to prune a subset of weights, such as filters connected to a channel.
Pruning methods using group lasso, which is one of the structured sparse regularization methods, were proposed by many researchers \cite{wen2016learning, alvarez2016learning, zhou2016less}. 
Mitsuno et al. \cite{mitsuno2020filter} proposed a pruning method to prune filters connected to a channels by using hierarchical group sparse regularization\cite{mitsuno2020hierarchical}.
Pruning of redundant filters with the scaling parameter of batch normalization layers was also proposed by enforcing sparsity of the parameters \cite{liu2017learning,ye2018rethinking,huang2018data}.
Li et al. \cite{li2016pruning} and He et al. \cite{he2018soft} proposed to use the norm of the weights. The filters with relatively low weight magnitudes are removed as redundant filters.
He et al. \cite{he2019filter} proposed to prune the most replaceable filters containing redundant information and the relatively less essential filters using the norm-based criterion.

\subsection{Knowledge Distillation}
Knowledge distillation can transfer the knowledge of DNNs with a large parameter (teacher networks) to smaller shallow networks (student networks).
Ba et al. \cite{ba2014deep} proposed to use L2 loss between the input vectors of the softmax activation function (logits) of the teacher network and the student network.
Hinton et al. \cite{hinton2015distilling} introduced to use the KL-divergence with a temperature parameter to make the softmax outputs of the teacher network and the softmax outputs (probability) of the student network similar.
Romero et al. \cite{romero2014fitnets} introduced to map the student hidden layer to the prediction of the teacher hidden layer.
Zhang et al. \cite{zhang2018deep} presented a deep mutual learning (DML) strategy where, rather than one-way transfer between a static pre-defined teacher network and a student network, an ensemble of students learn collaboratively and teach each other throughout the training process.

\subsection{Neural Architecture Search}
NAS automatically finds the optimal neural network structure.
Zoph et al. \cite{zoph2016neural} used a recurrent neural network as the controller to search the optimal neural network architecture in variable-length architecture space.
Zoph et al. \cite{zoph2018learning} proposed the NAS algorithm to search for an architectural building block on a small dataset, and then the block was transferred to a larger dataset.
This approach is quite flexible as it may be scaled in terms of computational cost and parameters to quickly address a variety of problems.
Pham et al. \cite{pham2018efficient} proposed an efficient neural architecture search method by searching for an optimal subgraph within a large computational graph.
Also, Cai et al. \cite{cai2018efficient} proposed an efficient architecture search method based on a reinforcement learning agent as the meta-controller.
Cai et al. \cite{cai2018proxylessnas} introduced ProxylessNAS that can directly learn neural network architectures on the target task and target hardware without any proxy.
Liu et al. \cite{liu2018darts} and Dong and Yang \cite{dong2019searching} proposed a gradient-based NAS approach, that represents the search space as a directed acyclic graph.
Real et al. \cite{real2019regularized} introduce the tournament selection evolutionary algorithm.
Wu et al. \cite{wu2019fbnet} presented a differentiable neural architecture search framework that optimizes over a layer-wise search space and represents the search space by a stochastic supernet.

\subsection{Incremental Training}
Incremental training algorithm is a dynamic configuration technique for DNNs that achieves energy-accuracy trade-offs in runtime by training a network incrementally as sub-networks.
Tann et al. \cite{tann2016runtime} proposed an incremental training algorithm in which the subsets of the weights in the network were incrementally trained by keeping the remaining weights trained in earlier steps.
Xun et al. \cite{xun2020incremental} proposed a dynamic DNNs using incremental training and group convolution pruning. In the dynamic DNNs, the channels of the convolution layer are divided into groups. At runtime, the following groups can be pruned for inference time/energy reduction or added back for accuracy recovery without model retraining.
Istrate et al. \cite{istrate2018incremental} proposed an incremental training method that partitions the original network into sub-networks, which are then gradually incorporated in the running network during the training process. 
Yu et al. \cite{yu2018slimmable} introduced slimmable neural networks, that permit instant and adaptive accuracy-efficiency trade-offs at runtime by training divided networks.

\section{Proposed Method}

\begin{figure*}[tbp]
\centerline{\includegraphics[scale=0.8]{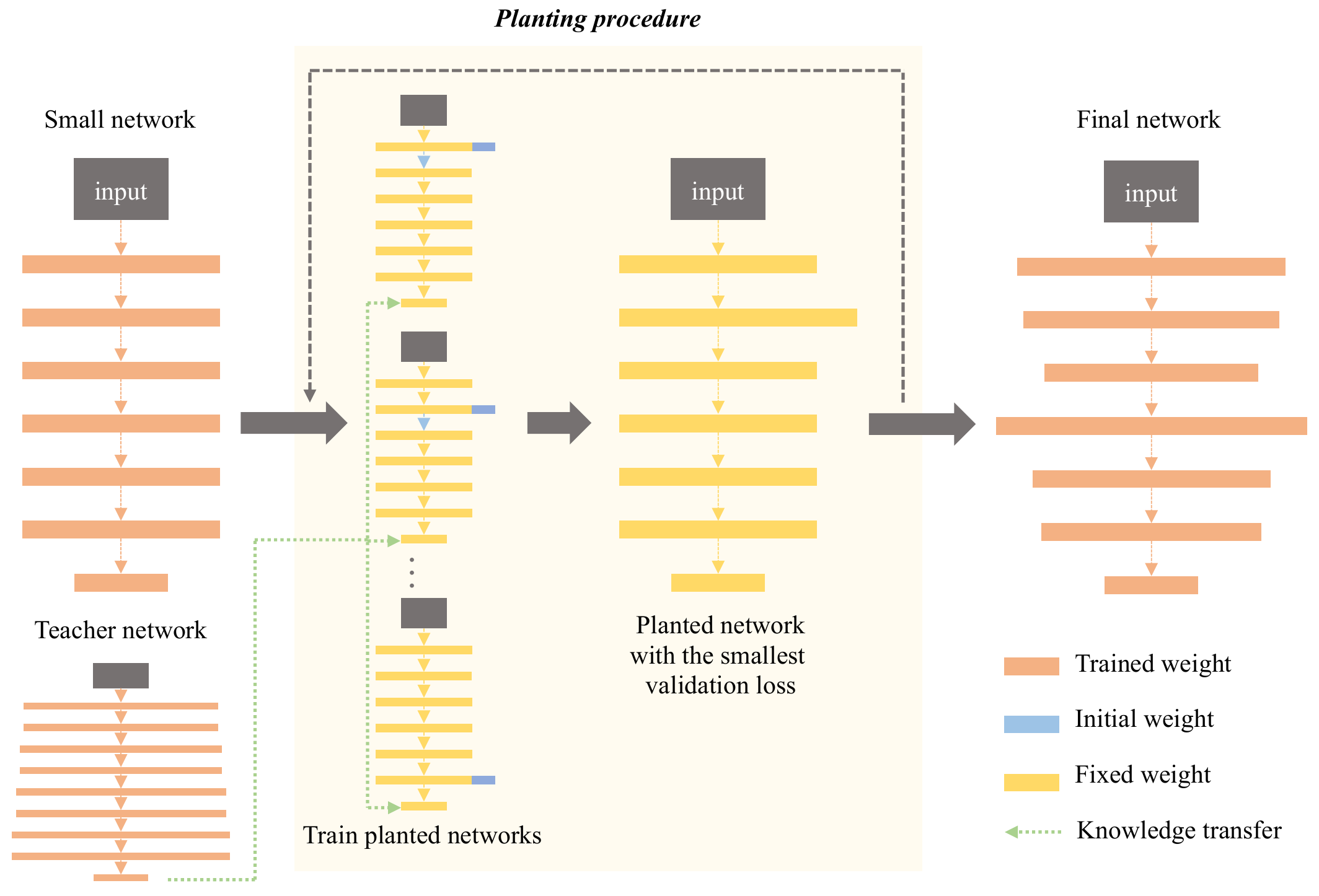}}
\caption{Illustration of Planting Procedure on a typical DNNs}
\label{Planting Procedure}
\end{figure*}

In this section, we propose a novel incremental training method for DNNs called {\em planting}. 
In the proposed incremental training method, channels on a small network are incrementally added to improve classification accuracy.
The parameters of the added channel are trained by using the knowledge distillation to imitate the behavior of the large network (the teacher network).
The optimal network architecture is searched by incrementally selecting the best channels among the possible candidates of the additions in terms of the classification accuracy (on the validation set).
The illustration of the proposed planting procedure on a typical DNN is shown in Fig. \ref{Planting Procedure}.

In summary, our planting approach consists of the following training processes: (0) training a large network as the teacher network.  (1) training a small network with fewer channels of each layer by a standard classification training method. (2) incrementally adding channels on the small network by using a knowledge distillation method with the teacher network.

\subsection{Planting Approach}

\textbf{Preparation}. 
We assume that the objective function of the optimization for determining the trainable weights is given by
\begin{equation}
\label{objective function}
J(W) = \mathbb{L}(f(x,W)|y)
\end{equation}
where $(x,y)$ denotes the pair of the input and target, $W$ is a set of all trainable weights of all the $L$ layers in the CNN,  $\mathbb{L}(\cdot)$ is the standard loss for the CNN.

Also, we assume the weight in the layer $l$ as $W^{l} \in \mathbb{R}^{C^{l}\times C^{l-1}\times K^{l} \times K^{l}}$, where $C^{l}$ and $C^{l-1}$ are the number of output channels and input channels, $K^{l}$ is the kernel size of the layer $l$ respectively. In the fully connected layers, $K^{l} = 1$.

\textbf{Initial network}. 
First, we train a small network with a few channels of each layer by a standard classification training procedure.
Also, we train a large network by a standard classification training procedure as the teacher network.
It is expected that the teacher network has the optimized number of channels with maximum performance in terms of classification accuracy.

\textbf{Search for the best layer for planting}. 
For considering the impact of planted channels on a network, we divide the layer of the small networks into several groups $G$, where $1 \le G \le L$. 
When using a very deep network, we plant channels in multiple layers by dividing a small network into groups to increase the impact of the planting.
Then, we add $n$ channels on the layer of group $g$. The number of channels of the added layer is given by 
\begin{equation}
C^{l}+n \;\; s.t. \;\; g*\frac{L}{G} \le l < (g+1)*\frac{L}{G}.\\
\end{equation}
The weights of the added channels of the small networks are trained 
by a knowledge distillation procedure while keeping the remaining weights are fixed.
The planted channels are learned to reduce the classification loss of the small network.
For example, when adding $n$ channels on the layer $l$, the weights $W^{l}_{C^{l}:C^{l}+n,:,:,:}$ and $W^{l+1}_{:,C^{l}:C^{l}+n,:,:}$ are trained.

We search the best group $g$ which minimizes the loss
\begin{equation}
\label{minimising loss}
\argmin_{g} J_{KL}(W^{S'},W^{L}),\\
\end{equation}
where $W^{S'}$ is the small network with the additional layer of group $g$ and 
$W^{L}$ is the large network. The detail definition of the loss $J_{KL}(W^{S'},W^{L})$ is explained in the next sub section.
The best layer to add is searched by using the brute-force search method or the random search method if there are many groups.
In the random search, some groups from G are randomly selected to reduce the calculation cost, and the best group is determined from the selected groups.
After we determined the best layer to reinforce, fix the planted channels and explore the next channel.
By repeating this planting process while reducing the classification loss than the previous network, we can obtained the best network architecture. 

After this method, we obtain a small network with fewer channels, which has higher performance than the networks obtained in a standard training procedure and can prevent over-fitting.
The network architecture is automatically optimized by the proposed planting procedures.
The details of the planting algorithm is shown in Algorithm \ref{Planting algorithm}.

\begin{algorithm}
 \caption{Planting algorithm}
 \label{Planting algorithm}
 \begin{algorithmic}[1]
 \renewcommand{\algorithmicrequire}{\textbf{Input:}}
 \REQUIRE $W^{S}$ : trained small network, $W^{L}$ : trained teacher network, $G$ : the number of group, $n$ : the number of planted channels, $(x_{train},y_{train})$ and $(x_{val},y_{val})$ : the training samples and the validation samples obtained by splitting the training set into two disjoint subsets.
 \WHILE{1}
  \FOR {$g$ in 1 \ldots $G$}
    \STATE $W^{S_{g}}=W^{S}$
    \FOR {$l$ in 1 \ldots $L$}
     \IF{$g*\frac{L}{G} \le l < (g+1)*\frac{L}{G}$}
        \STATE plant $n$ channels on layer $l$ of $W^{S_{g}}$
     \ENDIF
    \ENDFOR
    \STATE train $W^{S_{g}}$ via $J_{KL}(W^{S_{g}},W^{L})$ on $(x_{train},y_{train})$
  \ENDFOR
  \STATE $g_{min} = \argmin_{g\in G} |J(W^{S_{g}})|$ on $(x_{val},y_{val})$
  \IF{$J(W^{S_{g_{min}}}) \ge J(W^{S})$ on $(x_{val},y_{val})$}
    \STATE break
  \ENDIF
  \STATE $W^{S} = W^{S_{g_{min}}}$
  \ENDWHILE
 \end{algorithmic} 
 \end{algorithm}

\subsection{Knowledge distillation}
Knowledge distillation is an effective method for training the small network. 
In this study, we employ the Kullback Leibler (KL) Divergence. 
Suppose the predictions by the small network and the large network are $z^{S}$ and $z^{L}$ respectively, the KL divergence from $z^{S}$ to $z^{L}$ is given by
\begin{equation}
\mathbb{L_{KL}}(z^{L}||z^{S})
= \sum_{i} \frac{ \exp{z^{L}_{i}} }{ \sum_{j} \exp{z^{L}_{j}} } log \left( \frac{ \exp{z^{S}_{i}}}{ \sum_{j} \exp{z^{S}_{j}}} \right).
\end{equation}
The objective function for the proposed planting method is defined as follows
\begin{eqnarray}
J_{KL}(W^{S},W^{L}) = \lambda \mathbb{L}(f(x,W^{S})|y) \nonumber\\
+ ( 1 - \lambda ) \mathbb{ L_{KL}} (f(x,W^{L}) || f(x,W^{S}) ),
\end{eqnarray}
\rightline{ $s.t. \;\; 0\le \lambda \le 1$ \;\;\;\;  }
where $\lambda$ is used to balance the standard classification loss $\mathbb{L}(f(x,W^{S})|y) $ and KL divergence $\mathbb{L_{KL}}(f(x,W^{L})||f(x,W^{S}))$.

\section{Experiments}

To confirm the effectiveness of the proposed method, we have performed experiments with the image classification task using different datasets (CIFAR-10, CIFAR-100, and STL-10).

\section{Experiments using CIFAR-10}

\begin{table}[htbp]
\centering
\caption{The structure of networks}
\label{structure of networks}
\begin{tabular}{c|c}\hline
For CIFAR-10/100 & For STL-10 \\ \hline \hline
ReLU(conv1(kernel=3)) & ReLU(conv1(kernel=3)) \\
max pooling(2*2) & max pooling(2*2) \\
ReLU(conv2(kernel=3)) & ReLU(conv2(kernel=3)) \\
max pooling(2*2) & max pooling(2*2) \\
ReLU(conv3(kernel=3)) & ReLU(conv3(kernel=3)) \\
ReLU(conv4(kernel=3)) & max pooling(2*2) \\
ReLU(conv5(kernel=3)) & ReLU(conv4(kernel=3)) \\
max pooling(2*2) & ReLU(conv5(kernel=3)) \\
ReLU(fc1()) & max pooling(2*2) \\
output=fc2() & ReLU(fc1()) \\
 & output=fc2() \\
\hline
\end{tabular}
\end{table}

CIFAR-10 contains 60,000 color images of ten different animals and vehicles.
The size of each image is $32 \times 32$ pixels. 
They are divided into 45,000 training images, 5,000 validation images and 10,000 testing images.

In the experiments on CIFAR-10, we used the 7-layers CNN models with five convolutional layers and two fully connected layers, the structure of the network is shown in Table.\ref{structure of networks}.
All the experiments, we set the number of channels of the fully connected layers to $[ 128, 10]$.
All the number of channels of convolutional layers were set to $8$ for initial network and $128$ for the teacher network.

The initial network and the teacher network were trained from scratch by using SGD optimizer with a momentum of $0.9$.
We used the weight decay with the strength of $5*10^{-4}$ to prevent over-fitting.
The mini-batch size for CIFAR-10 was set to $128$ and the network was trained for $150$ epochs. 
The initial learning rate was set to $0.01$ and it was multiplied by $0.2$ after $[ 40, 80, 120]$ training epochs.

In the planting operation, we used the weight decay with the strength of $5*10^{-5}$, the number of the group $G$ was set to $5$, and other parameter settings are the same with the training of the initial network.
We added 4 channels to the layers at one planting operation. 
In the training of planted channels, the hyper-parameter $\lambda$ of KL loss (KLLoss) was set to $0$.
In the calculation for finding the smallest validation loss, the hyper-parameter $\lambda$ of KLLoss was set to $1$.

For comparing the performance of the proposed method, we trained the baseline networks with cross entropy loss (CELoss) as the standard classification loss, and KLLoss as loss function of knowledge transfer.
All the number of channels of the convolutional layer for the baseline networks were set to 8, 16, 32, 64 and 128, and we used the same teacher networks with the planting operation.
The hyper-parameter $\lambda$ of KLLoss was set to $0$.
Parameter settings are the same with the training of the initial network.

\begin{table}[htbp]
\caption{Results on CIFAR-10 dataset. The average of three trials are shown.}
\label{CIFAR-10 result}
\begin{tabular}{cc||cc||c}\hline
Network & Params & Test Err. & Test Acc. & Loss func\\ \hline \hline
\multirow{2}{*}{\begin{tabular}{c}Teacher$[128]$\\Student$[128]$\end{tabular}} & \multirow{2}{*}{\begin{tabular}{c}857.5K\end{tabular}} & 0.5007 & 88.10\% & CELoss\\ 
 & & 0.3823 & 88.51\% & KLLoss\\ \hline \hline 
\multirow{2}{*}{\begin{tabular}{c}Initial Network\\(Student$[8]$)\end{tabular}} & \multirow{2}{*}{\begin{tabular}{c}20.4K\end{tabular}} & 0.8300 & 71.55\% & CELoss\\ 
 & & 0.8245 & 71.69\% & KLLoss\\ \hline 
\multirow{2}{*}{\begin{tabular}{c}Student$[16]$\end{tabular}} & \multirow{2}{*}{\begin{tabular}{c}43.9K\end{tabular}} & 0.6071 & 79.42\% & CELoss\\ 
 & & 0.6108 & 79.23\% & KLLoss\\ \hline 
\multirow{2}{*}{\begin{tabular}{c}Student$[32]$\end{tabular}} & \multirow{2}{*}{\begin{tabular}{c}104.8K\end{tabular}} & 0.4898 & 84.03\% & CELoss\\ 
 & & 0.4791 & 84.02\% & KLLoss\\ \hline 
\multirow{2}{*}{\begin{tabular}{c}Student$[64]$\end{tabular}} & \multirow{2}{*}{\begin{tabular}{c}282.0K\end{tabular}} & 0.4431 & 86.83\% & CELoss\\ 
 & & 0.4103 & 86.80\% & KLLoss\\ \hline 
\hline Ours & 40.6K & 0.4825 & 84.35\% & KLLoss\\ 
\hline
\end{tabular}
\end{table}

The results for CIFAR-10 are shown in Table \ref{CIFAR-10 result}. In this table, the average of three trials are shown.
The number of channels of the convolutional layers after planting operation were $[ 12, 20, 16, 16, 12]$, $[ 12, 16, 16, 16, 16]$ and $[ 12, 16, 16, 16, 16]$.
For CIFAR-10, the proposed method is succeeded to train a network with higher classification accuracy, which has only $39\%$ parameters compare to a network where all the convolutional layers are 32 channels.

\section{Experiments using CIFAR-100}

CIFAR-100 contains 60,000 color images of 100 different categories.
The size of each image is $32 \times 32$ pixels. 
They are divided into 45,000 training images, 5,000 validation images and 10,000 testing images.

In the experiments on CIFAR-100, we used the same network structures with the experiments on CIFAR-10.
All the experiments, we set the number of channels of the fully connected layers to $[ 128, 100]$.
All the number of channels of convolutional layers were set to $16$ for the initial network and $128$ for the teacher network.
In the calculation for finding the smallest validation loss, the hyper-parameter $\lambda$ of KLLoss was set to $0$.
Other parameter settings are the same as the experiments on CIFAR-10.
For comparison of the performance of the proposed method with the standard methods, we trained the baseline networks on the settings of the same experiment with the experiments on CIFAR-10.

\begin{table}[htbp]
\caption{Results on CIFAR-100 dataset. The average of three trials are shown.}
\label{CIFAR-100 result}
\begin{tabular}{cc||cc||c}\hline
Network & Params & Test Err. & Test Acc. & Loss func\\ \hline \hline
\multirow{2}{*}{\begin{tabular}{c}Teacher$[128]$\\Student$[128]$\end{tabular}} & \multirow{2}{*}{\begin{tabular}{c}869.1K\end{tabular}} & 2.5010 & 57.76\% & CELoss\\ 
 & & 1.6232 & 60.05\% & KLLoss\\ \hline \hline 
\multirow{2}{*}{\begin{tabular}{c}Student$[8]$\end{tabular}} & \multirow{2}{*}{\begin{tabular}{c}32.0K\end{tabular}} & 2.5280 & 36.53\% & CELoss\\ 
 & & 2.5053 & 36.90\% & KLLoss\\ \hline 
\multirow{2}{*}{\begin{tabular}{c}Initial Network\\(Student$[16])$\end{tabular}} & \multirow{2}{*}{\begin{tabular}{c}55.5K\end{tabular}} & 2.1190 & 45.45\% & CELoss\\ 
 & & 2.0679 & 46.66\% & KLLoss\\ \hline 
\multirow{2}{*}{\begin{tabular}{c}Student$[32]$\end{tabular}} & \multirow{2}{*}{\begin{tabular}{c}116.5K\end{tabular}} & 1.9022 & 52.15\% & CELoss\\ 
 & & 1.7805 & 53.72\% & KLLoss\\ \hline 
\multirow{2}{*}{\begin{tabular}{c}Student$[64]$\end{tabular}} & \multirow{2}{*}{\begin{tabular}{c}293.6K\end{tabular}} & 1.9510 & 55.74\% & CELoss\\ 
 & & 1.6707 & 57.71\% & KLLoss\\ \hline 
\hline Ours & 78.5K & 1.7584 & 54.31\% & KLLoss\\ 
\hline
\end{tabular}
\end{table}

The results for CIFAR-100 are shown in Table \ref{CIFAR-100 result}. In this table, the average of three trials are shown.
The number of channels of the convolutional layers after planting operation were $[ 20, 24, 20, 24, 24]$, $[ 20, 24, 20, 24, 24]$ and $[ 20, 24, 24, 24, 20]$.
For CIFAR-100, the proposed method is succeeded to train a network with higher classification accuracy, which has only $67\%$ parameters compare to a network where all the convolutional layers are 32 channels.

\section{Experiments using STL-10}

STL-10 contains 13,000 color images of ten animals and vehicles.
The size of the image is $96 \times 96$ pixels.
They are divided into 5,000 training images, 1,000 validation images and 7,000 testing images.

In the experiments on STL-10, we used the 7-layers CNN models with five convolutional layers and two fully connected layers, the structure of the network is shown in Table.\ref{structure of networks}.
In all the experiments, we set the number of channels of the fully connected layers to $[ 128, 10]$.
All the number of channels of convolutional layers were set to $8$ for initial network and $64$ for the teacher network.

The network was trained for $100$ epochs, the initial learning rate was set to $0.01$ and it was multiplied by $0.1$ after every $epoch/3$ training epochs.
In the planting operation, we used the weight decay with the strength of $5*10^{-4}$.
In the calculation for finding the smallest validation loss, the hyper-parameter $\lambda$ of KLLoss was set to $0$.
Other parameter settings are the same with with the experiments on CIFAR-10.

For comparing the performance of the proposed method, we trained the baseline networks on the settings of the same experiment with the experiments on CIFAR-10.

\begin{table}[htbp]
\caption{Results on STL-10 dataset. The average of three trials are shown.}
\label{STL-10 result}
\begin{tabular}{cc||cc||c}\hline
Network & Params & Test Err. & Test Acc. & Loss func\\ \hline \hline
\multirow{2}{*}{\begin{tabular}{c}Teacher$[64]$\\Student$[64]$\end{tabular}} & \multirow{2}{*}{\begin{tabular}{c}445.8K\end{tabular}} & 1.5360 & 66.33\% & CELoss\\ 
 & & 1.1807 & 66.47\% & KLLoss\\ \hline \hline 
\multirow{2}{*}{\begin{tabular}{c}Initial Network\\(Student$[8]$)\end{tabular}} & \multirow{2}{*}{\begin{tabular}{c}40.8K\end{tabular}} & 1.2776 & 55.55\% & CELoss\\ 
 & & 1.2682 & 54.99\% & KLLoss\\ \hline 
\multirow{2}{*}{\begin{tabular}{c}Student$[16]$\end{tabular}} & \multirow{2}{*}{\begin{tabular}{c}84.9K\end{tabular}} & 1.2924 & 59.34\% & CELoss\\ 
 & & 1.1998 & 61.10\% & KLLoss\\ \hline 
\multirow{2}{*}{\begin{tabular}{c}Student$[32]$\end{tabular}} & \multirow{2}{*}{\begin{tabular}{c}186.8K\end{tabular}} & 1.2213 & 64.57\% & CELoss\\ 
 & & 1.1712 & 64.07\% & KLLoss\\ \hline 
\multirow{2}{*}{\begin{tabular}{c}Student$[128]$\end{tabular}} & \multirow{2}{*}{\begin{tabular}{c}1.2M\end{tabular}} & 1.7612 & 67.04\% & CELoss\\ 
 & & 1.1643 & 67.71\% & KLLoss\\ \hline 
\hline Ours & 82.6K & 1.0772 & 67.12\% & KLLoss\\ 
\hline
\end{tabular}
\end{table}

The results for STL-10 are shown in Table \ref{STL-10 result}. Again the average of three trials are shown in this table.
The number of channels of the convolutional layers after planting operation were $[ 28, 20, 20, 12, 12]$, $[ 28, 16, 20, 20, 16]$ and $[ 12, 20, 16, 28, 16]$.
For STL-10, the proposed method is succeeded to train a network with higher classification accuracy, which has only $7\%$ parameters compare to a network where all the convolutional layers are 128 channels with CELoss.
The test loss of planted network is the smallest than all the comparison networks and also the planting method can train to reduce over-fitting.



\section{Conclusion}

In this paper, we proposed a novel incremental training algorithm for deep neural networks called planting. 
Our planting approach can automatically search the optimal network architecture for training tasks with smaller parameters by planting channels incrementally to layers of the initial networks while keeping the earlier trained channels fixed for improving the network performances.
Also, we proposed to use the knowledge distillation method for training the channels planted. 
By transferring the knowledge of deeper and wider networks, we can grow the networks effectively and efficiently.
We evaluated the effectiveness of the proposed method on different datasets. 
We confirmed that the proposed approach was able to achieve comparable performance with smaller parameters compare to the larger network and reduce the over-fitting caused by a small amount of the data.

\bibliographystyle{unsrt}
\bibliography{references.bib}

\begin{thebibliography}{10}

\bibitem{krizhevsky2012imagenet}
Alex Krizhevsky, Ilya Sutskever, and Geoffrey~E Hinton.
\newblock Imagenet classification with deep convolutional neural networks.
\newblock In {\em Advances in neural information processing systems}, pages
  1097--1105, 2012.

\bibitem{szegedy2015going}
Christian Szegedy, Wei Liu, Yangqing Jia, Pierre Sermanet, Scott Reed, Dragomir
  Anguelov, Dumitru Erhan, Vincent Vanhoucke, and Andrew Rabinovich.
\newblock Going deeper with convolutions.
\newblock In {\em Proceedings of the IEEE conference on computer vision and
  pattern recognition}, pages 1--9, 2015.

\bibitem{he2016deep}
Kaiming He, Xiangyu Zhang, Shaoqing Ren, and Jian Sun.
\newblock Deep residual learning for image recognition.
\newblock In {\em Proceedings of the IEEE conference on computer vision and
  pattern recognition}, pages 770--778, 2016.

\bibitem{huang2017densely}
Gao Huang, Zhuang Liu, Laurens Van Der~Maaten, and Kilian~Q Weinberger.
\newblock Densely connected convolutional networks.
\newblock In {\em Proceedings of the IEEE conference on computer vision and
  pattern recognition}, pages 4700--4708, 2017.

\bibitem{lecun1990optimal}
Yann LeCun, John~S Denker, and Sara~A Solla.
\newblock Optimal brain damage.
\newblock In {\em Advances in neural information processing systems}, pages
  598--605, 1990.

\bibitem{hassibi1993second}
Babak Hassibi and David~G Stork.
\newblock Second order derivatives for network pruning: Optimal brain surgeon.
\newblock In {\em Advances in neural information processing systems}, pages
  164--171, 1993.

\bibitem{han2015learning}
Song Han, Jeff Pool, John Tran, and William Dally.
\newblock Learning both weights and connections for efficient neural network.
\newblock In {\em Advances in neural information processing systems}, pages
  1135--1143, 2015.

\bibitem{wen2016learning}
Wei Wen, Chunpeng Wu, Yandan Wang, Yiran Chen, and Hai Li.
\newblock Learning structured sparsity in deep neural networks.
\newblock In {\em Advances in neural information processing systems}, pages
  2074--2082, 2016.

\bibitem{mitsuno2020filter}
Kakeru Mitsuno and Takio Kurita.
\newblock Filter pruning using hierarchical group sparse regularization for
  deep convolutional neural networks.
\newblock submitted, 2020.

\bibitem{li2016pruning}
Hao Li, Asim Kadav, Igor Durdanovic, Hanan Samet, and Hans~Peter Graf.
\newblock Pruning filters for efficient convnets.
\newblock {\em arXiv preprint arXiv:1608.08710}, 2016.

\bibitem{he2018soft}
Yang He, Guoliang Kang, Xuanyi Dong, Yanwei Fu, and Yi~Yang.
\newblock Soft filter pruning for accelerating deep convolutional neural
  networks.
\newblock {\em arXiv preprint arXiv:1808.06866}, 2018.

\bibitem{liu2017learning}
Zhuang Liu, Jianguo Li, Zhiqiang Shen, Gao Huang, Shoumeng Yan, and Changshui
  Zhang.
\newblock Learning efficient convolutional networks through network slimming.
\newblock In {\em Proceedings of the IEEE International Conference on Computer
  Vision}, pages 2736--2744, 2017.

\bibitem{ba2014deep}
Jimmy Ba and Rich Caruana.
\newblock Do deep nets really need to be deep?
\newblock In {\em Advances in neural information processing systems}, pages
  2654--2662, 2014.

\bibitem{hinton2015distilling}
Geoffrey Hinton, Oriol Vinyals, and Jeffrey Dean.
\newblock Distilling the knowledge in a neural network.
\newblock In {\em NIPS Deep Learning and Representation Learning Workshop},
  2015.

\bibitem{zoph2016neural}
Barret Zoph and Quoc~V Le.
\newblock Neural architecture search with reinforcement learning.
\newblock {\em arXiv preprint arXiv:1611.01578}, 2016.

\bibitem{pham2018efficient}
Hieu Pham, Melody~Y Guan, Barret Zoph, Quoc~V Le, and Jeff Dean.
\newblock Efficient neural architecture search via parameter sharing.
\newblock {\em arXiv preprint arXiv:1802.03268}, 2018.

\bibitem{dong2019searching}
Xuanyi Dong and Yi~Yang.
\newblock Searching for a robust neural architecture in four gpu hours.
\newblock In {\em Proceedings of the IEEE Conference on Computer Vision and
  Pattern Recognition}, pages 1761--1770, 2019.

\bibitem{wu2019fbnet}
Bichen Wu, Xiaoliang Dai, Peizhao Zhang, Yanghan Wang, Fei Sun, Yiming Wu,
  Yuandong Tian, Peter Vajda, Yangqing Jia, and Kurt Keutzer.
\newblock Fbnet: Hardware-aware efficient convnet design via differentiable
  neural architecture search.
\newblock In {\em Proceedings of the IEEE Conference on Computer Vision and
  Pattern Recognition}, pages 10734--10742, 2019.

\bibitem{tann2016runtime}
Hokchhay Tann, Soheil Hashemi, R~Iris Bahar, and Sherief Reda.
\newblock Runtime configurable deep neural networks for energy-accuracy
  trade-off.
\newblock In {\em 2016 International Conference on Hardware/Software Codesign
  and System Synthesis (CODES+ ISSS)}, pages 1--10. IEEE, 2016.

\bibitem{molchanov2016pruning}
Pavlo Molchanov, Stephen Tyree, Tero Karras, Timo Aila, and Jan Kautz.
\newblock Pruning convolutional neural networks for resource efficient
  inference.
\newblock {\em arXiv preprint arXiv:1611.06440}, 2016.

\bibitem{lin2018accelerating}
Shaohui Lin, Rongrong Ji, Yuchao Li, Yongjian Wu, Feiyue Huang, and Baochang
  Zhang.
\newblock Accelerating convolutional networks via global \& dynamic filter
  pruning.
\newblock In {\em IJCAI}, pages 2425--2432, 2018.

\bibitem{peng2019collaborative}
Hanyu Peng, Jiaxiang Wu, Shifeng Chen, and Junzhou Huang.
\newblock Collaborative channel pruning for deep networks.
\newblock In {\em International Conference on Machine Learning}, pages
  5113--5122, 2019.

\bibitem{molchanov2019importance}
Pavlo Molchanov, Arun Mallya, Stephen Tyree, Iuri Frosio, and Jan Kautz.
\newblock Importance estimation for neural network pruning.
\newblock In {\em Proceedings of the IEEE Conference on Computer Vision and
  Pattern Recognition}, pages 11264--11272, 2019.

\bibitem{louizos2017learning}
Christos Louizos, Max Welling, and Diederik~P Kingma.
\newblock Learning sparse neural networks through $ l\_0 $ regularization.
\newblock {\em arXiv preprint arXiv:1712.01312}, 2017.

\bibitem{alvarez2016learning}
Jose~M Alvarez and Mathieu Salzmann.
\newblock Learning the number of neurons in deep networks.
\newblock In {\em Advances in Neural Information Processing Systems}, pages
  2270--2278, 2016.

\bibitem{zhou2016less}
Hao Zhou, Jose~M Alvarez, and Fatih Porikli.
\newblock Less is more: Towards compact cnns.
\newblock In {\em European Conference on Computer Vision}, pages 662--677.
  Springer, 2016.

\bibitem{mitsuno2020hierarchical}
Kakeru Mitsuno, Junichi Miyao, and Takio Kurita.
\newblock Hierarchical group sparse regularization for deep convolutional
  neural networks.
\newblock {\em arXiv preprint arXiv:2004.04394}, 2020.

\bibitem{ye2018rethinking}
Jianbo Ye, Xin Lu, Zhe Lin, and James~Z Wang.
\newblock Rethinking the smaller-norm-less-informative assumption in channel
  pruning of convolution layers.
\newblock {\em arXiv preprint arXiv:1802.00124}, 2018.

\bibitem{huang2018data}
Zehao Huang and Naiyan Wang.
\newblock Data-driven sparse structure selection for deep neural networks.
\newblock In {\em Proceedings of the European conference on computer vision
  (ECCV)}, pages 304--320, 2018.

\bibitem{he2019filter}
Yang He, Ping Liu, Ziwei Wang, Zhilan Hu, and Yi~Yang.
\newblock Filter pruning via geometric median for deep convolutional neural
  networks acceleration.
\newblock In {\em Proceedings of the IEEE Conference on Computer Vision and
  Pattern Recognition}, pages 4340--4349, 2019.

\bibitem{romero2014fitnets}
Adriana Romero, Nicolas Ballas, Samira~Ebrahimi Kahou, Antoine Chassang, Carlo
  Gatta, and Yoshua Bengio.
\newblock Fitnets: Hints for thin deep nets.
\newblock {\em arXiv preprint arXiv:1412.6550}, 2014.

\bibitem{zhang2018deep}
Ying Zhang, Tao Xiang, Timothy~M Hospedales, and Huchuan Lu.
\newblock Deep mutual learning.
\newblock In {\em Proceedings of the IEEE Conference on Computer Vision and
  Pattern Recognition}, pages 4320--4328, 2018.

\bibitem{zoph2018learning}
Barret Zoph, Vijay Vasudevan, Jonathon Shlens, and Quoc~V Le.
\newblock Learning transferable architectures for scalable image recognition.
\newblock In {\em Proceedings of the IEEE conference on computer vision and
  pattern recognition}, pages 8697--8710, 2018.

\bibitem{cai2018efficient}
Han Cai, Tianyao Chen, Weinan Zhang, Yong Yu, and Jun Wang.
\newblock Efficient architecture search by network transformation.
\newblock In {\em Thirty-Second AAAI conference on artificial intelligence},
  2018.

\bibitem{cai2018proxylessnas}
Han Cai, Ligeng Zhu, and Song Han.
\newblock Proxylessnas: Direct neural architecture search on target task and
  hardware.
\newblock {\em arXiv preprint arXiv:1812.00332}, 2018.

\bibitem{liu2018darts}
Hanxiao Liu, Karen Simonyan, and Yiming Yang.
\newblock Darts: Differentiable architecture search.
\newblock {\em arXiv preprint arXiv:1806.09055}, 2018.

\bibitem{real2019regularized}
Esteban Real, Alok Aggarwal, Yanping Huang, and Quoc~V Le.
\newblock Regularized evolution for image classifier architecture search.
\newblock In {\em Proceedings of the aaai conference on artificial
  intelligence}, volume~33, pages 4780--4789, 2019.

\bibitem{xun2020incremental}
Lei Xun, Long Tran-Thanh, Bashir Al-Hashimi, and Geoff Merrett.
\newblock Incremental training and group convolution pruning for runtime dnn
  performance scaling on heterogeneous embedded platforms.
\newblock In {\em ACM/IEEE Workshop on Machine Learning for CAD 2019
  (MLCAD'19)}, pages 1--6, 2020.

\bibitem{istrate2018incremental}
Roxana Istrate, Adelmo Cristiano~Innocenza Malossi, Costas Bekas, and Dimitrios
  Nikolopoulos.
\newblock Incremental training of deep convolutional neural networks.
\newblock {\em arXiv preprint arXiv:1803.10232}, 2018.

\bibitem{yu2018slimmable}
Jiahui Yu, Linjie Yang, Ning Xu, Jianchao Yang, and Thomas Huang.
\newblock Slimmable neural networks.
\newblock In {\em International Conference on Learning Representations}, 2019.

\end{thebibliography}

\end{document}